\documentclass[11pt]{article}

\usepackage{latex/acl}

\usepackage{times}
\usepackage{latexsym}
 \usepackage{booktabs} 
\usepackage{float}
 \usepackage{enumitem}
 \usepackage{amsmath, amssymb}
\usepackage[most]{tcolorbox}
\tcbset{
  colback=gray!5,
  colframe=black!60,
  boxrule=0.8pt,
  arc=3pt,
  outer arc=3pt,
  boxsep=5pt,
  left=6pt,
  right=6pt,
  top=6pt,
  bottom=6pt,
}

\usepackage[table]{xcolor} 

\usepackage[T1]{fontenc}
\usepackage{xcolor}
\usepackage{amsmath}
\usepackage[utf8]{inputenc}
\usepackage{microtype}

\usepackage{inconsolata}
\newcommand{\decomposetune}{\textsc{DecompTune}}

\usepackage{graphicx}

\title{Decomposition-Enhanced Training for Post-Hoc Attributions \\ in Language Models}

\author{
Sriram Balasubramanian$^{1,2}\thanks{$^*$Research work conducted during internship at Adobe Research}$, 
Samyadeep Basu$^{1}$, 
Koustava Goswami$^{1}$, 
Ryan Rossi$^{1}$, \\
\textbf{Varun Manjunatha$^{1}$, Roshan Santhosh$^{3}$, Ruiyi Zhang$^{1}$, 
Soheil Feizi$^{2}$, Nedim Lipka$^{1}$}\\
$^{1}$Adobe Research ,
$^{2}$University of Maryland, College Park,
$^{3}$Adobe Systems \\
\texttt{samyadeepb@adobe.com, sriramb@umd.edu}
}

\begin{document}
\maketitle
\begin{abstract}
Large language models (LLMs) are increasingly used for long-document question answering, where reliable attribution to sources is critical for trust. Existing post-hoc attribution methods work well for extractive QA but struggle in multi-hop, abstractive, and semi-extractive settings, where answers synthesize information across passages.
To address these challenges, we argue that post-hoc attribution can be reframed as a \textit{reasoning} problem, where answers are decomposed into constituent units, each tied to specific context. We first show that prompting models to generate such decompositions alongside attributions improves performance. Building on this, we introduce \decomposetune{}, a post-training method that teaches models to produce answer decompositions as intermediate reasoning steps. We curate a diverse dataset of complex QA tasks, annotated with decompositions by a strong LLM, and post-train Qwen-2.5 (7B and 14B) using a two-stage SFT + GRPO pipeline with task-specific curated rewards.
Across extensive experiments and ablations, \decomposetune{} substantially improves attribution quality, outperforming prior methods and matching or exceeding state-of-the-art frontier models.
\end{abstract}

\section{Introduction}
Large Language Models (LLMs) have demonstrated impressive capabilities across a range of natural language tasks, including question answering~\citep{kamalloo-etal-2023-evaluating}, summarization~\citep{liu2024learningsummarizelargelanguage}, and open-ended generation~\citep{openai2024gpt4technicalreport}. As their outputs increasingly influence downstream applications in real-world products such as AI Assistants~\citep{mialon2023gaiabenchmarkgeneralai, gabriel2024ethicsadvancedaiassistants}, the need for trustworthy and transparent generation becomes critical. One key aspect of this is attribution—the ability to cite supporting evidence from input sources that justifies each part of the model’s output. Although attributions can in principle be generated alongside answer generation, doing so often requires significant modifications to existing pipelines that already deliver strong answer-generation performance. To alleviate this problem, recent works~\citep{phukan2024peering, phukan2025logitlenscontextualembeddings, ramu2024enhancingposthocattributionslong} generate these attributions in a post-hoc manner i.e., given a document, question and a generated answer, they attribute the generated answer to relevant parts of the document. These post-hoc attribution methods are often based on various forms of semantic search~\citep{10.1561/1500000019} or identifying signals from the model internals (e.g., attention-heads or token embeddings)~\citep{hirsch2025laquerlocalizedattributionqueries}. These approaches perform well on purely extractive QA, but often
might underperform in more complex scenarios, particularly in multi-hop, semi-extractive and abstractive QA. In such cases, the surface-level answer often obscures the underlying reasoning steps or the multiple sources that were fused during generation. Without access to these intermediate computations, existing attribution methods frequently retrieve incomplete or irrelevant citations.

\begin{figure*}[t]
    \centering
    \includegraphics[width=\textwidth]{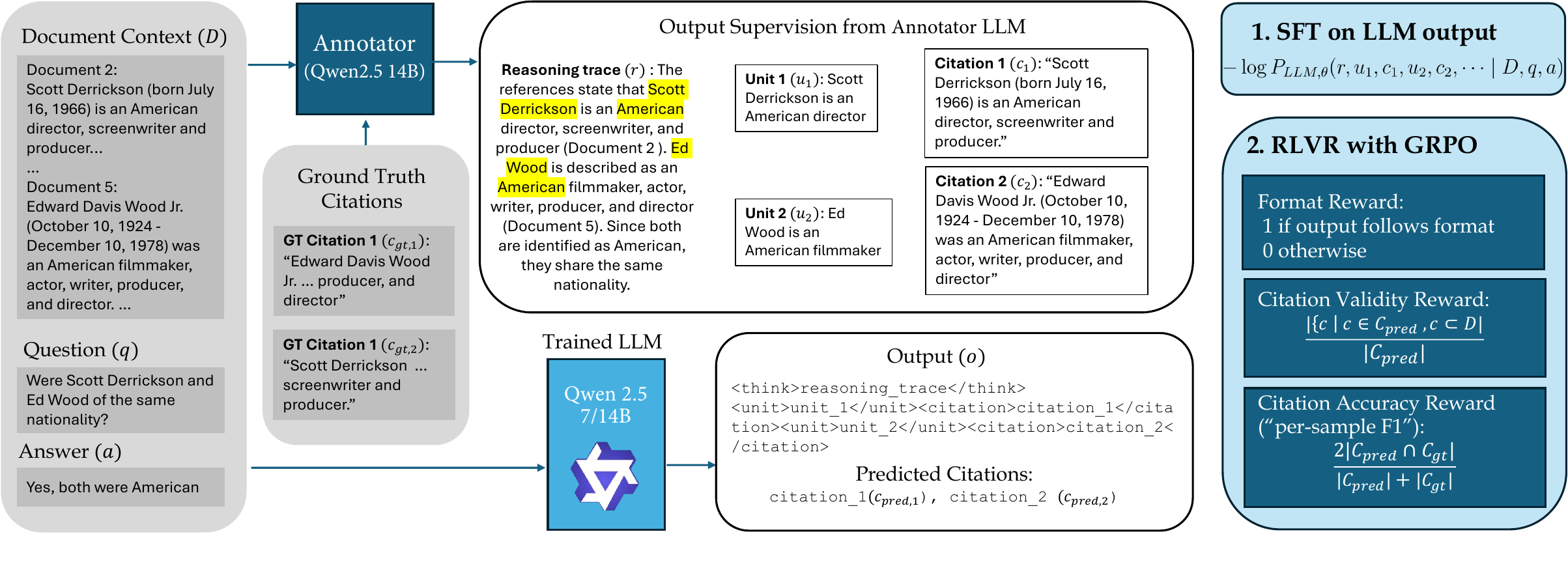}
    \caption{\textbf{Overview of our method \decomposetune{}. }We train small- to medium-sized language models to incorporate {\it answer decomposition as an intermediate reasoning step} prior to generating final attributions. To enable this, we leverage decomposition traces produced by an annotator LLM and adopt a two-stage training strategy: SFT to instill decomposition-aware reasoning, followed by GRPO with task-specific continuous and verifiable rewards to further refine attribution quality.}
    \label{fig:master_figure}
    \vspace{-0.3cm}
\end{figure*}

We argue that this limitation arises from a fundamental mismatch:  post-hoc attribution is treated as a \textit{retrieval} problem~\citep{phukan2024peering, hirsch2025laquerlocalizedattributionqueries}, when in fact it is often a \textit{reasoning} problem. For instance, a multi-hop question may require the model to infer intermediate facts (e.g., the birth country of a historical figure, followed by the laws of that country), each grounded in distinct pieces of evidence. Similarly, abstractive summaries often interleave paraphrased and synthesized content from several documents. In both settings, identifying appropriate citations necessitates reasoning by decomposing the model’s output answer into logical units that can each be grounded in specific source content. In our paper, we first verify that prompting a strong language model to generate these decompositions as an intermediate reasoning step and then generating the attributions together leads to strong improvements in attribution quality for cases involving multi-hop QA and complex semi-extractive QA setups.

Using this observation, we introduce \decomposetune, a post-training method and framework which trains small to intermediate sized language models to generate post-hoc attributions with a {\it reasoning-based decomposition step} as an intermediate step, rather than directly generating the attributions. To this end, we first curate a data-mixture encompassing various multi-hop, semi-extractive and abstractive QA where each QA pair is annotated with the intermediate decompositions using an annotator LLM (e.g. Qwen-2.5-14B). We then train Qwen-2.5-7B, and 14B using a two-stage sequential training approach. In the first stage, we train the language model on our data-mixture using SFT. In the second stage, we use GRPO with a set of continuous rewards (including format reward, accuracy reward and retrieval or validity reward). With this two stage training approach, we find that the post-trained Qwen series of models using \decomposetune~outperform existing open-source post-hoc attribution methods by a significant margin of at least 17 percentage points and also obtains a performance improvement of 23 points over the base model. Our post-trained models also obtain performance at-par with the best performing frontier models for most datasets and also outperform them in a few cases. 

Overall, in summary our paper makes the following contributions:

\begin{itemize}

\item We find that generating {\it answer decompositions as an intermediate reasoning step} improves attribution - all in one forward pass. 

\item  We curate a large-scale dataset mixture annotated with answer decompositions and their corresponding source attributions. This dataset serves as a valuable resource for post-training language models to perform attribution as a structured reasoning task.

\item We post-train small to intermediate sized language models using a two stage SFT + GRPO (with continuous rewards) training on our curated dataset, outperforming existing post-hoc attribution methods and obtaining performance at-par with closed-source models. 

\end{itemize}

\section{Related Works} \label{sec:related-work}

Attributing the output of a neural network to its input has been a long-standing problem in NLP \citep{Madsen_2022, rogers2020primerbertologyknowbert} and computer vision \citep{kazmierczak2025explainabilityvisionfoundationmodels, shrikumar2019learningimportantfeaturespropagating}, and has historically been approached from the lens of interpretability. In the pre-LLM era, the tasks that neural networks were capable of solving were small-scale, with input/output size at most a few paragraphs (in NLP) or images (in vision). The primary unit for attribution for these tasks were similarly small, with gradient-based and activation-based interpretability methods attributing the output to tokens or image patches. With the advent of powerful closed-source LLMs \citep{liu2025comprehensivesurveylongcontext} and new paradigms like multi-doc or long-context QA, several attribution methods \citep{nakano2022webgptbrowserassistedquestionansweringhuman, qin-etal-2023-webcpm, menick2022teachinglanguagemodelssupport, batista2025think} utilize the LLM itself for providing citations at a higher level, typically documents or paragraphs. Many of these works aim to prompt or train LLMs produce citations \textit{jointly with the answer} so that all factual sentences can be grounded in some external document. \citet{zhang2024longcite} finetune open-source LLMs (via SFT) on their curated dataset LongCite which consists of automatically generated long-context QA instances with precise sentence-level citations. \citet{chuang2025selfcite} propose SelfCite which is trained via RL with an ablation-based reward to cite sentences in the context. Companies like \cite{anthropic2025citations} and \cite{perplexityGettingStartedOverview} have also exposed LLM-generated citations API in their respective APIs. 

In \textbf{post-hoc attribution}, citations are produced after the answer is generated in a separate step. Prior work has focused on strategies for prompting an LLM to return citations given the context, question and answer \cite{ramu2024enhancingposthocattributionslong}, interpretability techniques to identify relevant circuits \cite{basu2025on}, internal layers~\citep{phukan2024peering, hirsch2025laquerlocalizedattributionqueries, Qi_2024} or information retrieval (IR) based approach~\citep{ramu2024enhancingposthocattributionslong} in which the sentence most relevant to the answer is retrieved from the documents in the context. However, these methods do not properly handle cases where multiple sources of information are synthesized into one answer part.  To the best of our knowledge, we are the first to propose a decomposition based prompting strategy to generate data and train LLMs on this data, outperforming other methods in this domain. \citet{ramu2024enhancingposthocattributionslong} also employs answer decompositions; however, these are treated as separate, independent steps for decomposition and attribution, rather than as part of an intermediate reasoning process, nor do they train LLMs using this decomposition.










\section{Initial Observation: Decomposition As an Intermediate Step Improves Attribution}
\label{sec:motivation}
\vspace{-0.1cm}
\begin{figure} 
  \centering
  \includegraphics[width=\linewidth]{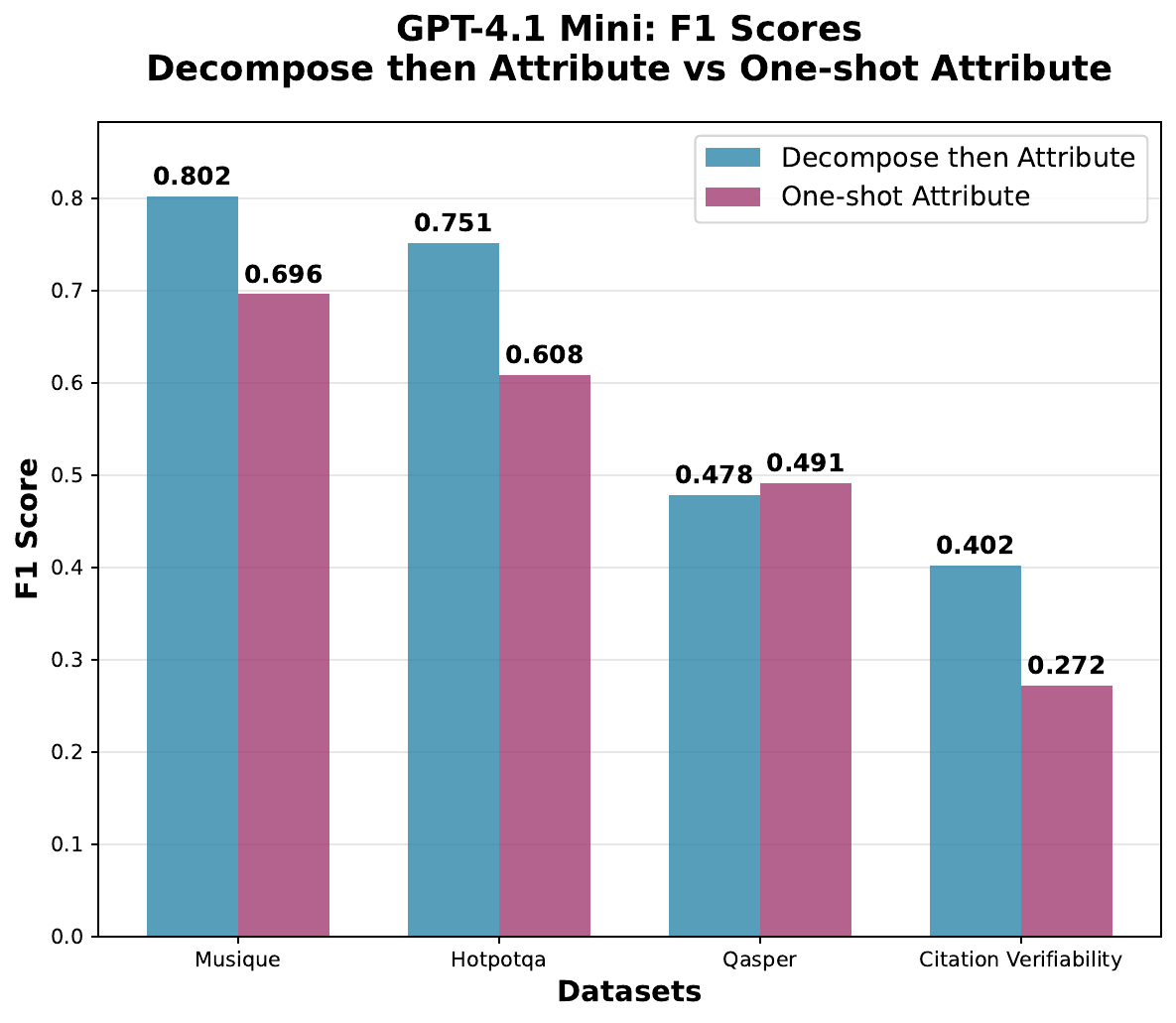}
  \vspace{-0.8cm}
  \caption{\textbf{Decomposition as intermediate step helps in attribution.} 
  Prompting LLMs to decompose an answer as an intermediate step, then attribute, 
  performs better than solely prompting them to only generate attributions, even with chain-of-thought prompting.}
  \label{fig:motivation}
    \vspace{-0.3cm}
\end{figure}
Earlier works~\citep{ramu2024enhancingposthocattributionslong} use answer decomposition as a pre-processing strategy where the decomposed answer units are then fed to an attributor separately (can be a LLM or a semantic search) for improved attribution.  
In our work, we prompt the language model to perform the decomposition step as an {\it intermediate reasoning step} and the attribution generation together. Given a document $D$, question $q$ and a generated answer $a$, we prompt an annotator language model to first decompose the answer $a$ into logical sub-units with a reasoning trace (see Fig.(\ref{fig:master_figure}) for an example of decomposition) and then generate the attributions for each of these sub-units. To improve the quality of the traces, we filter out sentences unrelated to the ground truth citations from the document context. 
From Fig.~\ref{fig:motivation}, we observe that prompting the language model to first generate the intermediate decomposition and then the final attributions yields substantial gains in attribution performance. Here, \textit{both} methods are instructed to think step-by-step and output their reasoning (exact DSPy \cite{khattab2024dspy} CoT prompts in Appendix~\ref{appendix:eval_prompts}), which implies that the difference in performance is solely due to decomposition into information units. In multi-hop and semi-extractive settings, relative performance improves by 13$\%$ and 12$\%$, respectively. These results confirm that introducing a decomposition step as an intermediate step prior to attribution generation, rather than producing attributions directly, leads to more accurate attributions. In the following sections, we describe the curation of training data and methods to train small-to-medium sized language models for post-hoc attribution using these answer decompositions as intermediate steps. 
\section{\decomposetune: Training Language Models to Decompose and Attribute}

Our method, in brief, consists of curating and processing open-source datasets covering our targeted domains (\ref{sec:data_construct}), and the details of the training algorithms employed for training the models (\ref{sec:training_details}).

\subsection{Training Dataset Curation and Processing} \label{sec:data_construct}

We construct our training dataset covering the (a) \textbf{multi-hop extractive} setting in which in answer is fairly short (only a few tokens) but is synthesized from multiple source citations; and (b) \textbf{abstractive} or \textbf{semi-extractive} settings wherein the answers are longer and may contain multiple distinct parts, each of which have to be disentangled and then attributed back to the context. We choose MuSiQue \citep{trivedi2022musique} and HotPotQA \citep{yang2018hotpotqa} as our multi-hop datasets, and  QASPER \citep{dasigi-etal-2021-dataset}, Verifiability Judgements \citep{liu-etal-2023-evaluating} and LongCite \citep{zhang2024longcite} for the semi-extractive and abstractive  setting.  However, not all of these datasets have sufficiently long context or sentence level ground truth attributions. We also require detailed reasoning traces containing the decompositions to finetune our models. We thus process the datasets in three stages:

\textbf{Step 1 - Context Extension}: Certain datasets (MuSiQue and HotpotQA) contain documents that are very short, often limited to just a few sentences or a single paragraph. In these cases, attribution may be too trivial. To create a more challenging benchmark, we  generate extended versions  of these documents (containing at least 500 words). To perform this extension any strong LLM (open-source or closed-source) can be used. The extended text must (a) include all original sentences verbatim, and (b) avoid restating the information from the original sentences elsewhere in the document. These constraints ensure that the ground-truth citations remain valid for the longer documents.


\textbf{Step 2 - Intermediate Decompositions}: Training our LLMs via supervised fine-tuning (SFT) requires high-quality reasoning traces that break down answers into smaller components, each of which can be attributed to a specific sentence in the source text. To generate these traces, we prompt the annotator LLM to first \textit{decompose} answers into logical sub-parts. We note that the decomposition step is agnostic to a specific model and any capable strong open-source or closed-source model can carry out this step. In our experiments, we find that using stronger models can improve performance after SFT, but these differences vanish after further finetuning with GRPO. Therefore, we can use relatively weaker annotator LLMs such as Qwen2.5 14B without any significant performance loss.  

\textbf{Step 3 - Attribution Matching}: Post intermediate decompositions of the answer sentence, we attribute each decomposed sub-unit to the relevant source sentence. To enhance precision, we provide the model with only the ground-truth citations as its source material during this process. We note that the annotator LLM is primarily responsible for matching the decomposed unit with the ground-truth citations (already present). In particular, the annotator LLM is not responsible for generating the citations. 

We provide the prompts used for each of the processing stages in the Appendix (Section \ref{sec:prompts}). We obtain a dataset of around 90,000 samples after this processing, each containing the relevant documents $D$, the question $q$ and full answer $a$, the answer units $\{ u_i \}$, the corresponding ground-truth citations $\{ c_i \}$  and LLM-generated reasoning trace containing the decomposition $r$.



\subsection{Training Details} \label{sec:training_details}

We adopt a two-stage training procedure: first, we fine-tune the LLM via SFT on the intermediate decompositions; then, we further train the SFT-tuned LLM via GRPO with specifically curated rewards for the task of post-hoc attribution. We explain the details of these two stages below:

\textbf{SFT on reasoning traces}: We first use direct supervision and fine-tune the LLM to generate the reasoning trace $r$, the decomposition units $\{u_i\}$ and the corresponding citations $\{c_i\}$ we obtained from the annotator LLM (Section~\ref{sec:data_construct}). Formally, the LLM is trained to generate $r$, $\{u_i\}$ and $\{c_i\}$ given documents ${D}$, question ${q}$, answer ${a}$ by minimizing the standard cross-entropy loss for next token prediction $ \mathbb{E} \left[ - \log P_{LLM, \theta} \left( r, u_1, c_1, , \dots u_n, c_n \mid {D}, {q}, {a},\right) \right]$, where $P_\theta$ represents the conditional likelihood of the data according to the LLM with weights $\theta$. We format the SFT completions containing the reasoning trace, units and citations as a single string $o$: \texttt{<think> reasoning trace </think> <unit> unit 1 </unit> <citation> citation 1 </citation> ... <unit> unit n </unit> <citation> citation n </citation>}

During this stage, the model learns to output its reasoning containing the decomposition and the citations in the required format, greatly reducing the computational effort in the GRPO stage. We find that simply prompting the base models is not very effective due to the relatively poor instruction following capabilities of these models, thus necessitating an SFT stage. The reasoning trace $r$ also serves as a rich, low-noise, signal in the SFT optimization process, thus preparing the model for RL training with only scalar rewards. 



\textbf{RL with continuous verifiable rewards}: After a round of SFT, we train the model further via GRPO \cite{shao2024deepseekmathpushinglimitsmathematical} with continuous verifiable rewards. We use a \textbf{format} reward, \textbf{validity} reward and a \textbf{weighted accuracy} reward for each sample. We describe each of them below:

\textbf{Format Reward:} We give a reward of 1 to the model for following the format required for parsing the reasoning, units, and citations from the output; and 0 if the output does not follow the format.


\textbf{Validity Reward:} Also, we give a reward if the citations output by the model are valid, that is, present in the document verbatim. The reward is the fraction of predicted citations ($C_{pred}$) that is a sub-string of the document context $D$. Formally, $ R_{\text{validity}} = \frac{|\{ c \mid c \in C_{pred} , c \subset D\}|}{|C_{pred}|} $. Here, we use $ c \subset D$ to denote that $c$ is a substring of $D$.

\textbf{Weighted Accuracy Reward:} Most importantly, we incentivize accurate citations by using a ``per-sample weighted'' F1 score as reward. Given predicted citations $C_{pred}$, ground-truth citations $C_{gt}$, and a recall weight $\alpha$, the accuracy reward is $R_{\text{accuracy}} = \frac{| C_{pred} \cap C_{gt}|}{(1 - \alpha) |C_{pred}| + \alpha |C_{gt}|}$. This can be interpreted as the weighted harmonic mean of the precision and recall.

All the above rewards are moderately sparse, and can drop to zero even with minor changes to the citations. We thus shape the rewards by utilizing a string similarity function to give non-zero reward when the citations are close but not exactly correct. Let $S(a, b)$ denote the similarity between two strings $a$ and $b$ as a number between $0$ and $1$ . Then, 

\begin{small}
\begin{align*}
R_{\text{format}}(o) &= \frac{\max_{s \subset o, s \text{ satisfies format}} |s|}{|o|}\\
R_{\text{validity}}(C_{pred}, D)  &= \frac{\sum_{c \in C_{pred}} \max_{s \subset D, |s| = |c|} S(c, s)}{|C_{pred}|} \\
R_{\text{accuracy}}(C_{pred}, C_{gt})  &=  \frac{ \max_{M} \sum_{(i, j) \in M} S(c_{pred, i}, c_{gt, j})}{(1 - \alpha) |C_{pred}| + \alpha |C_{gt}|} 
\end{align*}
\end{small}
where $M$ can be any bipartite matching between $C_{pred}$ and $C_{gt}$.

In practice, we use fuzzy string matching based on Levenshtein distance for calculating the similarity scores $S(a, b)$, which makes for efficient calculation of all these rewards. We also set $\alpha=0.75$ rather than the usual F1 score with $\alpha=0.5$, as this broadly improves recall while maintaining precision (see Section~\ref{sec:ablations} for details).





\definecolor{cGreen}{HTML}{B7D5A6} 
\section{Experiments and Results}
Using \decomposetune, we train two open-source language models: (i) Qwen2.5-7B-1M, and (ii) Qwen2.5-14B-1M~\citep{qwen2025qwen25technicalreport}. By training across these representative model sizes, we aim to assess the effectiveness and scalability of \decomposetune{} under different resource regimes.
We now provide a detailed description of the baseline methods as well as the adopted evaluation metrics.

\begin{table*}[t]
\centering
\resizebox{0.99\linewidth}{!}{
\begin{tabular}{p{3.3cm}*{4}{ccc}ccc}
\toprule
& \multicolumn{6}{c}{Multi-hop extractive} & \multicolumn{6}{c}{Abstractive/Semi-extractive} \\

& \multicolumn{3}{c}{MusiQue} & \multicolumn{3}{c}{HotpotQa} & \multicolumn{3}{c}{Qasper} & \multicolumn{3}{c}{Verifiability} & \multicolumn{3}{c}{Overall} \\
\cmidrule(lr){2-4} \cmidrule(lr){5-7} \cmidrule(lr){8-10} \cmidrule(lr){11-13} \cmidrule(lr){14-16}
Methods & P & R & F1 & P & R & F1 & P & R & F1 & P & R & F1 & P & R & F1 \\
\midrule

Frontier Models \\

\hspace{5pt}GPT-4o & .979 & .236 & .380 & .713 & .413 & .523 & .533 & .361 & .430 & .343 & .314 & .328 & .642 & .331 & .415 \\

\hspace{5pt}GPT-4.1-mini & .941 & .699 & .802 & .724 & .781 & .751 & .531 & .435 & .478 & .349 & .474 & .402 & .636 & .597 & .608\\ 

\hspace{5pt}Claude-Sonnet 4  &.964 & .351 & .514 & .869 & .413 & .560 & .611 & .355 & .449 & .491 & .474 & .483 & .734 & .398 & .501\\

\hspace{5pt}DeepSeek-R1 & .963 & .428 & .592 & .809 & .481 & .603 & .478 & .304 & .372 & .418 & .411 & .414 & .667 & .406 & .495\\
 \hspace{5pt}\textbf{Ours: 7B} & \cellcolor{cGreen} .885 & \cellcolor{cGreen} .716 & \cellcolor{cGreen} .792 & \cellcolor{cGreen} .834 & \cellcolor{cGreen} .815 & \cellcolor{cGreen} .825 & \cellcolor{cGreen} .455 & \cellcolor{cGreen} .468 & \cellcolor{cGreen} .462 & \cellcolor{cGreen} .500 & \cellcolor{cGreen} .484 & \cellcolor{cGreen} .492 & \cellcolor{cGreen} .669 & \cellcolor{cGreen} .621 & \cellcolor{cGreen} .642  (+\textbf{.034}) \\ 
\hline
Other Methods\\
\hspace{5pt}Qwen2.5-3B * & .746 & .215 & .334 & .486 & .315 & .382 & .174 & .143 & .157 & .119 & .204 & .150 & .381 & .219 & .256\\

\hspace{5pt}Qwen2.5-7B * &  .869 & .393 & .542 & .587 & .482 & .529 & .282 & .324 & .302 & .209 & .334 & .257 & .487 & .383 & .407 \\
\hspace{5pt}Qwen2.5-14B *  &  .934 & .376 & .537 & .695 & .520 & .595 & .415 & .354 & .382 & .364 & .387 & .375 & .602 & .410 & .472\\
\hspace{5pt}Llama3.1-8B * & .832 & .563 & .671 & .534 & .571 & .552 & .178 & .258 & .211 & .137 & .404 & .204 & .420 & .449 & .409 \\

\hspace{5pt}Embeddings &  .711 & .263 & .384 & .631 & .282 & .390 & .391 & .285 & .330 & .381 & .360 & .370  & .520 & .297 & .370\\

\hspace{5pt}Attention-Head &   .753 & .282 & .410 & .612 & .297 & .400 & .425 & .291 & .345 & .393 & .362 & .377 & .545 & .308 & .394\\

 \hspace{5pt}\textbf{Ours:7B} & \cellcolor{cGreen} .885 & \cellcolor{cGreen} .716 & \cellcolor{cGreen} .792 & \cellcolor{cGreen} .834 & \cellcolor{cGreen} .815 & \cellcolor{cGreen} .825 & \cellcolor{cGreen} .455 & \cellcolor{cGreen} .468 & \cellcolor{cGreen} .462 & \cellcolor{cGreen} .500 & \cellcolor{cGreen} .484 & \cellcolor{cGreen} .492 & \cellcolor{cGreen} .669 & \cellcolor{cGreen} .621 & \cellcolor{cGreen} .642  (\textbf{+.170}) \\ 




\hline
\textbf{Ours: 7B}\\
 \hspace{5pt}SFT &  .848 &  .273 &  .414 &  .779 &  .445 &  .566 &  .516 &  .239 &  .326 &  .405 &  .249 &  .308 & \ .637 & \ .301 & \ .404 \\

\hspace{5pt}SFT + GRPO {\tiny ($\alpha=0.5$)}  & .904 & .709 & .795 & .835 & .776 & .805 & .564 & .351 & .433 & .525 & .380 & .441 & .707 & .554 & .618 \\
  \hspace{5pt}SFT + GRPO & \cellcolor{cGreen} .885 & \cellcolor{cGreen} .716 & \cellcolor{cGreen} .792 & \cellcolor{cGreen} .834 & \cellcolor{cGreen} .815 & \cellcolor{cGreen} .825 & \cellcolor{cGreen} .455 & \cellcolor{cGreen} .468 & \cellcolor{cGreen} .462 & \cellcolor{cGreen} .500 & \cellcolor{cGreen} .484 & \cellcolor{cGreen} .492 & \cellcolor{cGreen} .669 & \cellcolor{cGreen} .621 & \cellcolor{cGreen} .642 (\textbf{+.238}) \\

\hline
\textbf{Ours: 14B}\\
\hspace{5pt}SFT & .888 & .303 & .452 & .819 & .463 & .592 & .547 & .270 & .362 & .458 & .279 & .346 & .678 & .329 & .438 \\

\hspace{5pt}SFT + GRPO & \cellcolor{cGreen} .918 & \cellcolor{cGreen} .743 & \cellcolor{cGreen} .821 & \cellcolor{cGreen} .861 & \cellcolor{cGreen} .764 & \cellcolor{cGreen} .809 & \cellcolor{cGreen} .518 & \cellcolor{cGreen} .406 & \cellcolor{cGreen} .455 & \cellcolor{cGreen} .529 & \cellcolor{cGreen} .513 & \cellcolor{cGreen} .520 & \cellcolor{cGreen} .706 & \cellcolor{cGreen} .606 & \cellcolor{cGreen}{\color{blue} \textbf{.650}} (\textbf{+.212})\\

\bottomrule
\end{tabular}
}

\caption{\label{table:exact_math} \textbf{Evaluation Results using exact string match with ground-truth attributions.} We find that with exact-string matches to ground-truth attributions as an evaluation - our post-trained models surpasses frontier models and also obtain gains over other post-hoc attribution methods by a strong margin. We find that post-training with our curated GRPO rewards is very crucial in obtaining the best performance. The best performing open-source variant is marked with {\color{blue}Blue}. Improvement over the best performing method in each section is shown in parenthesis (Note that asterisked methods use a more involved prompting strategy with multiple LLM calls; see Section~\ref{sec:baselines} and Appendix~\ref{appendix:eval_prompts})}
\end{table*}

\subsection{Baselines}
\label{sec:baselines}
We compare our approach against three broad families of baselines, chosen to represent both strong language models and established post-hoc attribution methods. (i) {\it Frontier LLMs}. We include GPT-4o, GPT-4.1-mini, Claude-Sonnet 4 and DeepSeek-R1 as representative state-of-the-art frontier models. These models have consistently demonstrated strong performance across a wide range of reasoning and QA tasks, and therefore provide a competitive upper bound for attribution quality in closed settings. We use the decomposition-based prompting strategy we used in Section~\ref{sec:motivation}.
(ii) {\it Smaller LLMs}. To establish comparisons with smaller LLMs comparable to , we also evaluate against Qwen2.5-3B, Qwen2.5-7B-1M, Qwen2.5-14B-1M~\citep{qwen2025qwen25technicalreport} and Llama 3.1 8B~\citep{grattafiori2024llama3herdmodels}. This selection allows us to cover models of different parameter scales and training regimes, providing insight into how attribution performance varies across model size and capacity in publicly available systems. For these models, however, directly prompting them to decompose then attribute yields poor results because of inadequate instruction following capabilities. Therefore, we use a more involved strategy — first prompt to decompose into individual units, then in separate LLM calls, prompt to attribute each unit to a sentence in the reference — for much better performance across the board (see Appendix~\ref{appendix:eval_prompts} for exact prompts). However, this involves many more LLM calls per sample, with much larger inference time overhead making it less practical for deployment.

Finally, we evaluate two white-box attribution approaches: (a)  Token Embeddings~\citep{phukan2024peering}, which leverage token-level representation alignment; and (b) AttnAttribute~\citep{basu2025on} which uses attribution heads. For AttnAttribute, we follow the original formulation but adapt it by providing the generated answer as input within the assistant tag, along with the question and document context, and then extract attributions using the attention heatmap from the attribution head.

\subsection{Evaluations}
We evaluate our trained models on the test partitions of MusiQue~\citep{trivedi2022musique}, HotPotQA~\citep{yang2018hotpotqa}, Qasper~\citep{dasigi-etal-2021-dataset}, Verifiability~\citep{liu-etal-2023-evaluating}, as well as on the GovReport~\citep{huang-etal-2021-efficient} and arXiv-summarization datasets~\citep{cohan-etal-2018-discourse}. This collection of benchmarks spans a diverse range of evaluation settings, covering multi-hop reasoning, semi-extractive and abstractive question answering, and long-document summarization. Among these, MusiQue, HotPotQA, Qasper, and Verifiability provide ground-truth citations which enables us to measure attribution quality via an exact string match. We report the \textbf{precision}, \textbf{recall}, and \textbf{F1 score} for these datasets. The precision is the fraction of output citations that match the ground truth, the recall is the fraction of ground truth citations that are returned by the model, and the F1 score is the harmonic mean of the precision and recall.

\begin{table*}[t]
\centering
 \resizebox{0.99\linewidth}{!}{
\begin{tabular}{p{3.3cm}*{6}{cc}cc}
\toprule
& \multicolumn{4}{c}{Multi-hop extractive} & \multicolumn{4}{c}{Abstractive/Semi-extractive} & \multicolumn{4}{c}{Summarization} \\

& \multicolumn{2}{c}{MusiQue} & \multicolumn{2}{c}{HotpotQa} & \multicolumn{2}{c}{Qasper} & \multicolumn{2}{c}{Verifiability} & \multicolumn{2}{c}{GovReport} & \multicolumn{2}{c}{Arxiv-Summary} & \multicolumn{2}{c}{Overall} \\
\cmidrule(lr){2-3} \cmidrule(lr){4-5} \cmidrule(lr){6-7} \cmidrule(lr){8-9} \cmidrule(lr){10-11} \cmidrule(lr){12-13} \cmidrule(lr){14-15}
Methods & IS & CS & IS & CS & IS & CS & IS & CS & IS & CS & IS & CS & IS & CS \\
\midrule
\midrule

Frontier Models \\
\hspace{5pt}GPT-4o &.432 & .347 & .571 & .545 & .736 & .723 & .691 & .620 & .825 & .660 & .733 & .440 & .665 & .556 \\

\hspace{5pt}GPT-4.1-mini & .616 & .616 & .776 & .846 & .715 & .756 & .819 & .812 & .812 & .698 & .733 & .597 & .745 & .721 \\

\hspace{5pt}Claude-Sonnet 4  & .679 & .455 & .800 & .588 & .813 & .779 & .932 & .862 & .887 & .780 & .801 & .687 & .819 & .692 \\

\hspace{5pt}DeepSeek-R1 & .673 & .509 & .817 & .680 & .762 & .707 & .892 & .810 & .848 & .733 & .722 & .580 & .786 & .670 \\
\hspace{5pt} \textbf{Ours: 7B}  & \cellcolor{cGreen} .632 & \cellcolor{cGreen} .583 & \cellcolor{cGreen} .821 & \cellcolor{cGreen} .839 & \cellcolor{cGreen} .666 & \cellcolor{cGreen} .768 & \cellcolor{cGreen} .887 & \cellcolor{cGreen} .839 & \cellcolor{cGreen} .699 & \cellcolor{cGreen} .716 & \cellcolor{cGreen} .563 & \cellcolor{cGreen} .583 & \cellcolor{cGreen} .712 (-.107) & \cellcolor{cGreen} .721 (+0)  \\
\hline
Other Methods\\
\hspace{5pt}Qwen2.5-3B * & .320 & .229 & .541 & .390 & .396 & .331 & .614 & .462 & .476 & .555 & .336 & .265 & .447 & .372 \\

\hspace{5pt}Qwen2.5-7B * & .511 & .393 & .715 & .578 & .674 & .623 & .769 & .680 & .589 & .614 & .456 & .506 & .619 & .565 \\
\hspace{5pt}Qwen2.5-14B * & .571 & .423 & .790 & .653 & .741 & .671 & .892 & .724 & .758 & .566 & .668 & .425 & .737 & .577 \\
\hspace{5pt}Llama3.1-8B * & .461 & .486 & .630 & .650 & .346 & .465 & .590 & .689 & .342 & .741 & .307 & .286 & .446 & .553 \\




\hspace{5pt}Embeddings & .531 & .341 & .789 & .531 & .685 & .651 & .812 & .689 & .610 & .620 & .468 & .524 & .649 & .559\\ 

\hspace{5pt}Attention-Head &  .546 & .330 & .810 & .546 & .683 & .677 & .831 & .699 & .604 & .628 & .488 & .531 
& .660 & .568 \\ 
\hspace{5pt}\textbf{Ours: 7B}  & \cellcolor{cGreen} .632 & \cellcolor{cGreen} .583 & \cellcolor{cGreen} .821 & \cellcolor{cGreen} .839 & \cellcolor{cGreen} .666 & \cellcolor{cGreen} .768 & \cellcolor{cGreen} .887 & \cellcolor{cGreen} .839 & \cellcolor{cGreen} .699 & \cellcolor{cGreen} .716 & \cellcolor{cGreen} .563 & \cellcolor{cGreen} .583 & \cellcolor{cGreen} .712 (-.016) & \cellcolor{cGreen} .721 (\textbf{+.144}) \\
\hline




\textbf{Ours:7B}\\

\hspace{5pt}Only SFT & .468 & .344 & .731 & .596 & .580 & .521 & .709 & .545 & .717 & .390 & .570 & .237 & .629 & .439 \\

\hspace{5pt}SFT + GRPO {\tiny ($\alpha=0.5$)} & .611 & .547 & .819 & .818 & .659 & .647 & .837 & .705 & .734 & .489 & .602 & .381 & .710 & .598 \\

\hspace{5pt}SFT + GRPO & \cellcolor{cGreen} .632 & \cellcolor{cGreen} .583 & \cellcolor{cGreen} .821 & \cellcolor{cGreen} .839 & \cellcolor{cGreen} .666 & \cellcolor{cGreen} .768 & \cellcolor{cGreen} .887 & \cellcolor{cGreen} .839 & \cellcolor{cGreen} .699 & \cellcolor{cGreen} .716 & \cellcolor{cGreen} .563 & \cellcolor{cGreen} .583 & \cellcolor{cGreen} .712 (\textbf{+.083}) & \cellcolor{cGreen} .721 (\textbf{+.273}) \\

\hline
\textbf{Ours: 14B}\\

\hspace{5pt}Only SFT & .517 & .377 & .770 & .623 & .666 & .607 & .763 & .616 & .767 & .438 & .596 & .252 & .680 & .485 \\

\hspace{5pt}SFT + GRPO  & \cellcolor{cGreen}.620 & \cellcolor{cGreen} .569 & \cellcolor{cGreen} .825 & \cellcolor{cGreen} .837 & \cellcolor{cGreen} .688 & \cellcolor{cGreen} .771 & \cellcolor{cGreen} .891 & \cellcolor{cGreen} .873 & \cellcolor{cGreen} .768 & \cellcolor{cGreen} .730 & \cellcolor{cGreen} .649 & \cellcolor{cGreen} .651 & \cellcolor{cGreen} {\color{blue}\textbf{.736} }(\textbf{+.056}) & \cellcolor{cGreen}{\color{blue}\textbf{.738}} (\textbf{+.253})  \\ 
\bottomrule
\end{tabular}}

\caption{\label{table:llm_judge} \textbf{Evaluation Results with LLM-as-a-Judge}. We find that our post-trained small language models closely match frontier models and outperform existing post-hoc attribution methods. Even in this evaluation setup, we find that post-training with our curated GRPO rewards is very crucial in obtaining the best performance. The best performing open-source variant is marked with {\color{blue}Blue}.}
\end{table*}

However, not all benchmarks (GovReport and arxiv-summarization) contain such reference annotations. For consistency, we further adopt an \textbf{LLM-as-a-judge framework} to evaluate the attributions. This approach allows us to evaluate and compare the methods over all benchmarks in a unified manner. Given a question, answer, and a list of output citations, we prompt GPT 4.1 to give scores according to two rubrics: (a) \textbf{collective support score (CS)}: how well does the list of citations (taken collectively) support the answer (analogous to recall), and (b) \textbf{individual support score (IS)}: does each individual citation support any part of the answer to the question (analogous to precision). We report the average CS and IS over the entire dataset after normalizing the scores between 0 to 1. The exact prompt template employed in our LLM-as-a-judge evaluations is in the appendix (see Appendix \ref{appendix:llm_judge}). While a few methods (such as AttnAttribute) always output citations that are single sentences present in the document, LLM-powered methods may not always obey this condition. In such cases, we may encounter invalid or multi-sentence citations. Thus, we first filter out the citations that cannot be found in the context, and then break each of them into individual sentences if necessary. As these steps can be feasibly performed during inference time, we may view them as part of the methods themselves.

\section{Main Results}
\subsection{Improvements Over Strong Baselines} 
We find that our proposed approach, \decomposetune{}, yields substantial improvements over a wide range of baseline models. The SFT stage alone improves instruction following to the extent that attribution becomes possible in a single LLM call, with performance close to the more involved prompting strategy. In particular, our two-stage training framework, which combines supervised fine-tuning (SFT) with GRPO using our task-specific curated rewards, consistently enhances attribution quality across different evaluation settings. Specifically, we observe improvements in F1 score of upto 30 percentage points for multi-hop QA attribution tasks, 23 points for semi-extractive and abstractive attribution tasks, and upto 11 points of improvement in IS (and 23 points in CS) for summarization attribution tasks over the original models. These gains are especially notable for models in the 7B parameter range, where resource constraints typically limit performance.
Overall, our findings demonstrate that decomposition-enhanced training provides a reliable and generalizable recipe for strengthening attribution in small- to medium-scale models. This highlights the potential of carefully designed training strategies to unlock grounding capabilities in open-source language models.
\begin{figure*}[t]
    \centering
    \includegraphics[width=0.48\linewidth]{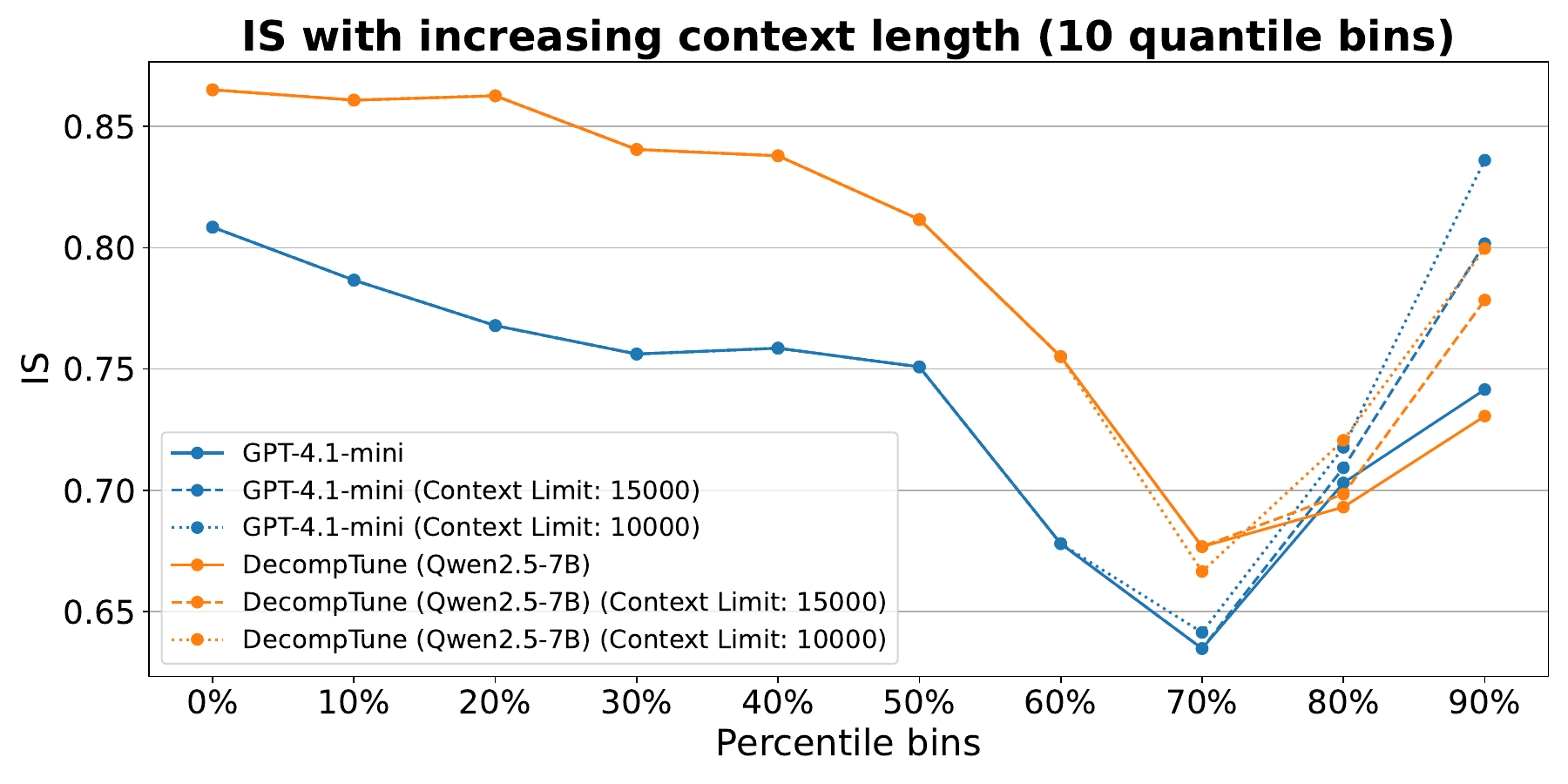}
     \includegraphics[width=0.48\linewidth]{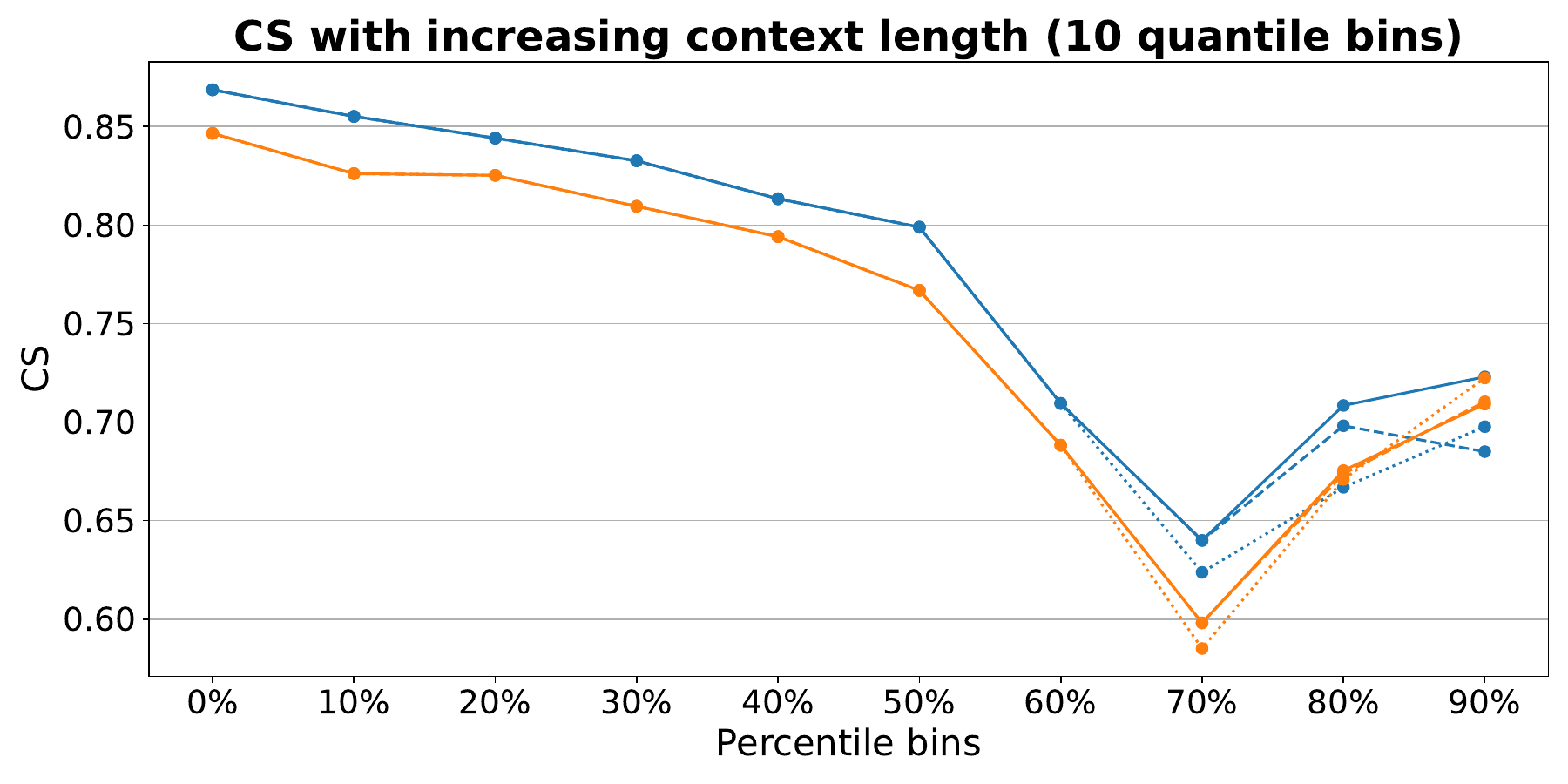}
    \vspace{-0.2cm}
    \caption{ \textbf{Effect of Context-Length on Post-Hoc Attribution}. IS and CS score on samples from all datasets of different lengths, grouped into 10 equal quantile bins. ~\decomposetune{} (7B) outperforms GPT 4.1 mini on IS, and is closely matched on CS. When coupled with BM25 based search to reduce context, all models gain significantly on IS, but CS for GPT-4.1 drops while ~\decomposetune{} improves. }
    \label{fig:acc_vs_context}
    \vspace{-0.3cm}
\end{figure*}
We also benchmark against Token-Embeddings and AttnAttribute, both of which are strong baselines that leverage white-box access to language model internals. These methods represent a more challenging comparison point, as they are specifically designed to exploit internal model signals for attribution. Nevertheless, we find that our trained models outperform both by a significant margin of at least 28 points, highlighting the robustness of decomposition-enhanced training. 
\vspace{-0.2cm}
\subsection{Comparisons with Frontier Models}
We also compare the performance of our trained models against strong frontier models. In particular, following our initial observations in Fig.~\ref{fig:master_figure}, we prompt the frontier models to first perform answer decomposition as an intermediate reasoning step before generating the final attributions. To ensure a fair comparison, we apply the same prompting strategy to both our trained models and the frontier systems.
As shown in Table~\ref{table:exact_math} and Table~\ref{table:llm_judge}, our trained models achieve performance that is on par with, and in some cases even exceeds, that of the frontier baselines. These results demonstrate not only the effectiveness of decomposition-enhanced training, but also its potential to narrow the gap between open-source and frontier models in the task of post-hoc attribution. Overall, our results show that our training strategy leads to strong attribution performance with 7B and 14B sized models, outperforms existing post-hoc attribution methods and obtains performance at-par with frontier models, providing an efficient alternative.  
\vspace{-0.2cm}
\begin{tcolorbox}
\textbf{Main Takeaway}: For post-hoc attribution, SFT alone yields {\it instruction following improvements}, whereas applying GRPO with task-specific curated rewards yields \textit{more substantial gains} for attribution.
\end{tcolorbox}

\section{Ablations and Further Analysis} \label{sec:ablations}
\begin{figure}
    \centering
    \includegraphics[width=0.99\linewidth]{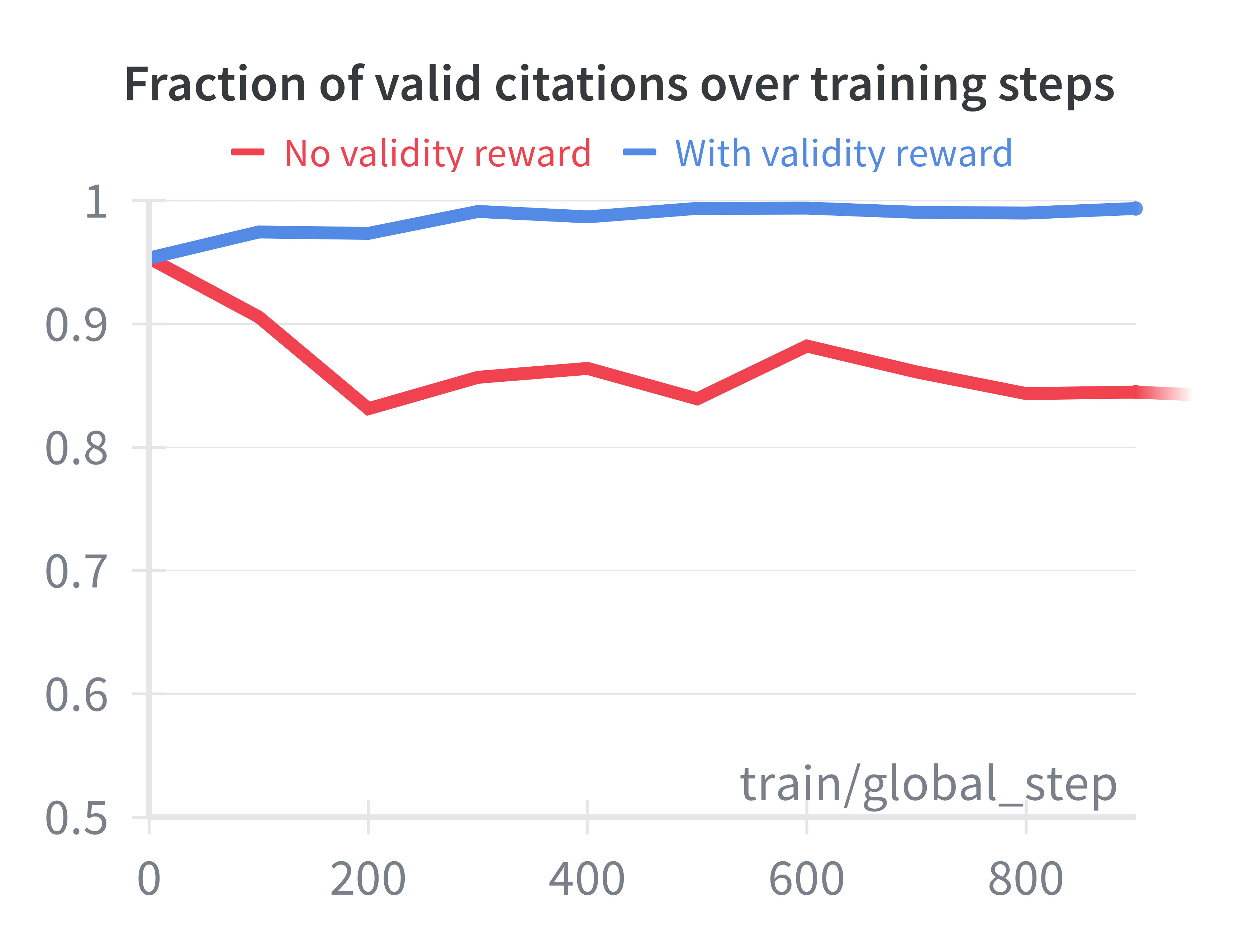}
    \vspace{-0.8cm}
    \caption{\textbf{Validity reward} ensures that the output citations are present verbatim in the context.}
    \label{fig:ablation_1}
    \vspace{-0.3cm}
\end{figure}
\textbf{Ablation on rewards:} We ablate the validity and recall-weighted accuracy reward to show their utility in performance of the model. In Figure~\ref{fig:ablation_1}, we plot the fraction of citations that are \textit{valid} (present verbatim in the document) as the RL training progresses. While initially high, the fraction quickly drops in the absence of a validity reward. With such a reward, however, the fraction of valid sentences is approximately 1. We also show the overall performance (IS, CS, precision, recall, and F1) of models trained with standard F1 score ($\alpha=0.5$) and trained with recall weighted F1 score ($\alpha=0.75$) in Table~\ref{table:exact_math} and Table~\ref{table:llm_judge}. The recall and collective support scores, which measure the extent to which the answer is supported by the citations, improve dramatically, while maintaining precision and IS. 

\textbf{Error analysis on input context length:} 
We analyze model performance across varying context lengths to assess how accuracy depends on reference size. Since very long documents exceed the context window, we also test whether performance holds when reducing context using a lightweight BM25-based sentence selection. Figure~\ref{fig:acc_vs_context} shows IS and CS for GPT-4.1-mini and ~\decomposetune{} (7B) across 10 context-length quantiles, as well as when contexts are capped at 10k and 15k characters. ~\decomposetune{} (7B) achieves notably higher IS than GPT-4.1-mini on short contexts and matches it on the top quantiles, though it slightly lags in CS across all bins. With BM25-based search, all models show strong IS gains on long contexts; however, GPT-4.1-mini’s CS declines, while ~\decomposetune{} maintains or improves it. We thus believe LLM-based attribution methods can scale well to longer contexts when coupled with search.

\vspace{-0.2cm}
\section{Conclusion}
\vspace{-0.2cm}
In this paper, we introduced \decomposetune{}, to train small models for post-hoc attribution. Instead of generating attributions directly from answers, \decomposetune{} guides models to decompose answers into constituent units that can be more reliably aligned with supporting evidence. Our training pipeline combines SFT for instruction-following followed by GRPO with task-specific rewards. Across diverse benchmarks, \decomposetune{} consistently outperforms existing methods and matches/surpasses frontier model performance.


\section{Limitations}
While our work introduces a new approach to attribution in post-training language models, it does not pose direct ethical risks. On the contrary, attribution methods have the potential to mitigate harms by grounding model outputs in identifiable sources and making the generative process more interpretable. By connecting generated content back to inputs, our approach can support accountability, reduce misinformation, and provide users with greater transparency into model behavior. Nonetheless, ethical challenges in this area remain broader than our scope: attribution alone cannot fully address issues such as dataset bias, privacy, or intellectual property. We emphasize, however, that our contribution is directed toward enhancing interpretability and grounding, and therefore serves as a positive step toward more trustworthy language model deployment.
\bibliography{latex/acl_latex}

\appendix

\section{Prompts For Dataset Construction} \label{sec:prompts}

\begin{tcolorbox}[title=Prompt to extend context, fonttitle=\bfseries,fontupper=\small]

Seamlessly extend this passage by around \texttt{\{num\_words\}} words:

\medskip
\texttt{\{passage\}}

\medskip
\textbf{Instructions:}
\begin{enumerate}[leftmargin=*, itemsep=4pt]
    \item In your output, you must include ALL the sentences in the original passage.
    \item Instead of repeating the sentence word-to-word, just insert the corresponding tag instead. For example, the output can be 
          ``This is a new sentence. \texttt{<s0>} This is another new sentence. \texttt{<s1>} And yet another one.''
    \item The original sentences should be scattered throughout the output, they should not appear in one place.
    \item VERY IMPORTANT: The information in the original sentences (in the \texttt{<si>} tags) should not be repeated in the scattered sentences, they should appear only once. In other words, you should NOT be able to infer the information in the original sentences by reading the rest of the document.
    \item Ensure that ALL tags \texttt{<s0>}, \texttt{<s1>}, \ldots, \texttt{<sn>} are present in the output, corresponding to the original sentences.
    \item Your output should JUST be the extended passage, with no additional text or formatting.
    \item Again, ensure that you DO NOT GENERATE any sentences that contain the information in the original sentences! Simply use the tags \texttt{<s0>}, \texttt{<s1>}, \ldots, \texttt{<sn>} to refer to the original sentences.
\end{enumerate}

\end{tcolorbox}

\begin{tcolorbox}[title=Prompt to generate sentence level attributions for Musique, fonttitle=\bfseries, fontupper=\small]

You are an expert at extracting sentence attributions from a context based on a question and answer.
You are given the following:
\medskip

Context:
\texttt{\{context\}}
\medskip

Question:
\texttt{\{question\}}
\medskip

Answer:
\texttt{\{answer\}}
\medskip

Your task is to extract the EXACT sentences from the context that are used for producing the answer to the question. Each sentence is separated by a newline in the context. 

You should return the sentences separated by newlines, with no additional text or formatting, in the order they appear in the context.

To derive the answer from the question, you will need at least one sentence from each paragraph in the context.

Do not include any sentences that are not necessary for answering the question.

VERY IMPORTANT: You should include at least one sentence from each paragraph in the context.

\end{tcolorbox}

\begin{tcolorbox}[title=Prompt to generate decompositions, fonttitle=\bfseries, width=0.5\textwidth, fontupper=\small]

You will be given a set of input references, a question, an \texttt{accumulated\_text}, and a \texttt{latest\_text\_chunk}.  
You need to break down the \texttt{latest\_text\_chunk} for a given question into information units. Give only those information units that are attributable to the documents.

\medskip
\textbf{Instruction on what good information units are:}
\begin{enumerate}[leftmargin=*, itemsep=4pt]
    \item Give information units that are relevant to the answer.
    \item Information units should be meaningful.
    \item All information units should be necessary for the answer to derived from the question and documents.
    \item The list of information units should be complete, the answer should be derivable solely from the listed information units.
    \item Each information unit should be attributable to a SINGLE sentence in the references. It should be worded in a form such that it can be easily searched in the references.
    \item Some extra information may need to be added to the information units so that they can be easily attributed to the references. For example, references may contain the sentences ``The Great Gatsby is a famous 1925 novel written by F. Scott Fitzgerald'' and ``The Great Gatsby was narrated by Nick Carraway.'', but the question and answer may not have the name of the novel (example: ``Who wrote the famous 1925 novel narrated by Nick Carraway? F. Scott Fitzgerald.''). In this case, the information units should contain the name of the novel too (``The Great Gatsby'') so that it can be attributed easily to the references.
    \item The decomposition may be trivial, that is, the \texttt{latest\_text\_chunk} may already be a well-defined information unit.
\end{enumerate}

\medskip
\textbf{Instruction on what bad information units are:}
\begin{enumerate}[leftmargin=*, itemsep=4pt]
    \item Information units that convey duplicate information.
    \item Information units that are non statements.
    \item Information units that are not meaningful to the question.
    \item Information units that are not attributable to the references.
    \item Information units that are not atomic, i.e, combine information from multiple sentences in the references.
\end{enumerate}

\end{tcolorbox}

\section{Judge LLM Prompt}
\label{appendix:llm_judge}

\begin{tcolorbox}[title=Prompt to score output citations, fonttitle=\bfseries, width=0.5\textwidth, fontupper=\tiny]

You are an expert judge. Given a question, a part of the answer, and a list of citations, determine whether the citations SUPPORT the answer.
Output consists of two parts, collective support score and individual support score.

Collective support score is defined as follows for the entire citation list:
\begin{enumerate}[leftmargin=*, itemsep=4pt]
\item If the citations FULLY support the answer part, then 2.
\item If the citations only PARTIALLY support the answer part, then 1.
\item If the citations do NOT support the answer part at all, then 0.
\end{enumerate}

Individual support score is defined as follows for each citation:
If the citation supports (fully or partially) the answer part, then 1.
If the citation does NOT support the answer part at all, then 0.

\medskip
\textbf{Examples}:

Question: What is the capital of the largest country in the world (by area)?

Answer part: The capital of the largest country, Russia, is Moscow.

Citation list:
[``Moscow is the capital of Russia.", ``Russia is the largest country in the world by area."]

Output: ``collective\_support\_score": 2, ``individual\_support\_scores": [1, 1]

Explanation: The citation list fully supports the answer part because together they confirm that Moscow is the capital of Russia and that Russia is the largest country by area. Individually, both citations support the answer part.

If citation list is:
[``Moscow is the capital of Russia."]

Output: ``collective\_support\_score": 1, ``individual\_support\_scores": [1]

Explanation: The citation list partially supports the answer part because while it confirms that Moscow is the capital of Russia, it does not provide information about Russia being the largest country by area. Individually, the single citation supports the answer part.

If citations are:
[``The capital of France is Paris.", ``The largest country by area is Russia.", ``Moscow is the capital of Russia."]

Output:
{{
    ``collective\_support\_score": 2,
    ``individual\_support\_scores": [0, 1, 1]
}}

Explanation: The citation list fully supports the answer part because two of the citations confirm that Moscow is the capital of Russia and that Russia is the largest country by area. Individually, the first citation does not support the answer part, while the other two do.

\medskip

Now evaluate the following:
\medskip

Question: \texttt{\{question\}}
\medskip

Answer part: \texttt{\{answer\_part\}}
\medskip

Citation list (this is what you are evaluating):
\texttt{\{citation\_list\}}
\medskip

\textbf{Instructions:}
\begin{enumerate}[leftmargin=*, itemsep=4pt]
\item Ensure that you do not use any information or knowledge outside of the snippet when evaluating. 

\item If the citation list is empty, \texttt{collective\_support\_score} should be 2 if there is no factual information in the answer part, else 0. \texttt{individual\_support\_scores} should be an empty list. 

\item Your output should be in the following JSON format without any additional text, no backticks or code blocks.
\end{enumerate}
\texttt{\{ "collective\_support\_score": 0, 1, or 2,  "individual\_support\_scores": [list of 0s and 1s corresponding to each citation]\}}

\end{tcolorbox}

\section{Prompts For Evaluation}
\label{appendix:eval_prompts}

\newpage 
\begin{tcolorbox}[title={DSPY Prompt to decompose, then attribute in the same call (with CoT)}, fonttitle=\bfseries, width=0.5\textwidth, fontupper=\small]

You will be given a set of input references, a question, an \texttt{accumulated\_text}, and a \texttt{latest\_text\_chunk}.
You need to break down the \texttt{latest\_text\_chunk} for a given question into information units. Then for each unit, find ONE citation from the references that supports it. Each citation should be verbatim from the references.
You should output a dictionary where each key is an information unit and the value is a citation from the references that supports it.


\textbf{Instruction on what good information units are}:

\begin{enumerate}[leftmargin=*, itemsep=4pt]

\item Give information units that are relevant and meaningful to the answer.
\item All information units should be necessary for the answer to derived from the question and documents.
\item The list of information units should be complete, the answer should be derivable solely from the listed information units.
\item Each information unit should be attributable to a SINGLE sentence in the references. It should be worded in a form such that it can be easily searched in the references.
\item Some extra information may need to be added to the information units so that they can be easily attributed to the references. For example, references may contain the sentences "The Great Gatsby is a famous 1925 novel written by F. Scott Fitzgerald" and "The Great Gatsby was narrated by Nick Carraway.", but the question and answer may not have the name of the novel (example: "Who wrote the famous 1925 novel narrated by Nick Carraway? F. Scott Fitzgerald."). In this case, the information units should contain the name of the novel too ("The Great Gatsby") so that it can be attributed easily to the references.
\item The decomposition may be trivial, that is, the \texttt{latest\_text\_chunk} may already be a well-defined information unit.

\end{enumerate}


\textbf{Instruction on what bad information units are:}

\begin{enumerate}[leftmargin=*, itemsep=4pt]

\item Information units that convey duplicate information.
2. Information units that are non statements.
3. Information units that are not meaningful to the question.
4. Information units that are not attributable to the references.
5. Information units that are not atomic, i.e, combine information from multiple sentences in the references.

\end{enumerate}

\medskip 

\textbf{Format instructions:}

\begin{itemize}[leftmargin=*, itemsep=4pt]
\item Remember to add \#\# at the end of each field name in the output, the format of `\#\# \texttt{<field\_name>} \#\#' is IMPORTANT. 
\item The name of each field in the output should EXACTLY match the name in the instructions.
\end{itemize}

\end{tcolorbox}

\newpage 






\newpage
\newpage 
\begin{tcolorbox}[title={DSPY Prompt to only decompose into units (with CoT)}, fonttitle=\bfseries, fontupper=\small]

You will be given a set of input references, a question, an \texttt{accumulated\_text}, and a \texttt{latest\_text\_chunk}.
You need to break down the \texttt{latest\_text\_chunk} for a given question into information units. Give only those information units that are attributable to the documents.

\textbf{Instruction on what good information units are}:

\begin{enumerate}[leftmargin=*, itemsep=4pt]

\item Give information units that are relevant and meaningful to the answer.
\item All information units should be necessary for the answer to derived from the question and documents.
\item The list of information units should be complete, the answer should be derivable solely from the listed information units.
\item Each information unit should be attributable to a SINGLE sentence in the references. It should be worded in a form such that it can be easily searched in the references.
\item Some extra information may need to be added to the information units so that they can be easily attributed to the references. For example, references may contain the sentences "The Great Gatsby is a famous 1925 novel written by F. Scott Fitzgerald" and "The Great Gatsby was narrated by Nick Carraway.", but the question and answer may not have the name of the novel (example: "Who wrote the famous 1925 novel narrated by Nick Carraway? F. Scott Fitzgerald."). In this case, the information units should contain the name of the novel too ("The Great Gatsby") so that it can be attributed easily to the references.
\item The decomposition may be trivial, that is, the \texttt{latest\_text\_chunk} may already be a well-defined information unit.

\end{enumerate}


\textbf{Instruction on what bad information units are:}

\begin{enumerate}[leftmargin=*, itemsep=4pt]

\item Information units that convey duplicate information.
2. Information units that are non statements.
3. Information units that are not meaningful to the question.
4. Information units that are not attributable to the references.
5. Information units that are not atomic, i.e, combine information from multiple sentences in the references.

\end{enumerate}

\textbf{Format instructions:}

\begin{itemize}[leftmargin=*, itemsep=4pt]
\item Remember to add \#\# at the end of each field name in the output, the format of `\#\# \texttt{<field\_name>} \#\#' is IMPORTANT. 
\item The name of each field in the output should EXACTLY match the name in the instructions.
\end{itemize}

\end{tcolorbox}

\begin{tcolorbox}[title={DSPY Prompt to attribute each unit}, fonttitle=\bfseries, fontupper=\small]

You will be given a set of input references and a query sentence. Your job is to find the sentence in the references that supports the query most closely and output it verbatim.

\medskip 

\textbf{Follow these rules:}

\begin{enumerate}[leftmargin=*, itemsep=4pt]
\item The output sentence should be only a SINGLE SENTENCE from the references and should match EXACTLY.
\item DO NOT output multiple sentences.
\end{enumerate}

\end{tcolorbox}

\begin{tcolorbox}[title={Prompt used for training in ~\decomposetune{}}, fonttitle=\bfseries, fontupper=\small]

You need to produce attributions to ground an answer to a question based on relevant documents. The inputs are:
\medskip

\#\# \textbf{Question} \#\#

\texttt{\{question\}}

\medskip
\#\# \textbf{Relevant documents} \#\#

\texttt{\{documents\}}

\medskip

\#\# \textbf{Answer generated till now} \#\#

\texttt{\{answer\_till\_now\}}

\medskip

\#\# \textbf{Answer part} \#\#

\texttt{\{answer\_part\}}

\medskip

You need to ground the answer part to the relevant documents by first decomposing the answer part into reasoning, constituent units and citations.

\end{tcolorbox}

\newpage
\begin{tcolorbox}[title={DSPY Prompt to directly attribute (with CoT)}, fonttitle=\bfseries, fontupper=\small]

You will be given a set of input references, a question, an \texttt{accumulated\_text}, and a \texttt{latest\_text\_chunk}.
You need to output all the citations that support the \texttt{latest\_text\_chunk}. Each citation should be verbatim from the references.
Each citation should be necessary to support the \texttt{latest\_text\_chunk}. All citations collectively should be sufficient to support the \texttt{latest\_text\_chunk}.

\medskip 

\textbf{Format instructions:}

\begin{itemize}[leftmargin=*, itemsep=4pt]
\item Remember to add \#\# at the end of each field name in the output, the format of `\#\# \texttt{<field\_name>} \#\#' is IMPORTANT. 
\item The name of each field in the output should EXACTLY match the name in the instructions.
\end{itemize}

\end{tcolorbox}
\newpage 

\section{Hyperparameters for Training}

We describe the hyperparameters for the SFT and GRPO training stages below:

\textbf{SFT}: For all models, we use a batch size of 64 packed sequences of length 32768 tokens and learning rate is $4 \times 10^{-5}$ with Adam optimizer. We select the best model based on the validation loss on a held out validation dataset.

\textbf{GRPO}: For all models, we use a batch size of 288 with group size 8. The maximum prompt length was 16384 with maximum completion length as 2048 tokens. All rewards were weighted equally, and the recall weight ($\alpha$) was 0.75. We select the best model based on the validation accuracy reward on a held out validation dataset. We use the corrected GRPO algorithm with no length or difficulty bias. Rollouts were generated using the vLLM engine.

All experiments were performed using at most 2 nodes of 8 A100 GPUs.

\section{Hyperparameters for AttnAttribute and Token-Embeddings}
\label{appendix: hyperparam_baselines}
For adapting the AttnAttribute method from~\citep{basu2025on}, we use the Qwen-2.5-7B-Instruct model with the attention layer, head of [15,30] which we find to be the attribution head. We use the same algorithm for span selection as in~\citep{basu2025on}. For Token-Embeddings from~\citep{phukan2024peering}, we use the same layer of 15th in Qwen-2.5-7B-Instruct, but use the distance between the hidden representations to select the relevant span above a high-threshold (we use 0.7 in our experiments).

\end{document}